\documentclass[journal]{journal}
\usepackage{ifpdf}
\usepackage{booktabs}
 \usepackage{multirow}

\usepackage{cite}
\ifCLASSINFOpdf
  \usepackage[pdftex]{graphicx}
\else
\fi
\usepackage{array}

\usepackage{fixltx2e}
\begin{document}
%
% paper title
% can use linebreaks \\ within to get better formatting as desired
\title{Linguistic features for sentence difficulty prediction in ABSA }
%
%
% author names and IEEE memberships
% note positions of commas and nonbreaking spaces ( ~ ) LaTeX will not break
% a structure at a ~ so this keeps an author's name from being broken across
% two lines.
% use \thanks{} to gain access to the first footnote area
% a separate \thanks must be used for each paragraph as LaTeX2e's \thanks
% was not built to handle multiple paragraphs
%

\author{Adrian-Gabriel~Chifu, Sébastien~Fournier
             % <-this % stops a space
\thanks{A.-G. Chifu and S. Fournier are with Aix-Marseille Université, Université de Toulon CNRS, LIS, Marseille, France e-mail: adrian.chifu@univ-amu.fr and sebastien.fournier@univ-amu.fr}}% <-this % stops a space

\maketitle
\thispagestyle{empty}

\begin{abstract}
%\boldmath
One of the challenges of natural language understanding is to deal with the subjectivity of sentences, which may express opinions and emotions that add layers of complexity and nuance. Sentiment analysis is a field that aims to extract and analyze these subjective elements from text, and it can be applied at different levels of granularity, such as document, paragraph, sentence, or aspect. Aspect-based sentiment analysis is a well-studied topic with many available data sets and models. However, there is no clear definition of what makes a sentence difficult for aspect-based sentiment analysis. In this paper, we explore this question by conducting an experiment with three data sets: "Laptops", "Restaurants", and "MTSC" (Multi-Target-dependent Sentiment Classification), and a merged version of these three datasets. We study the impact of domain diversity and syntactic diversity on difficulty. We use a combination of classifiers to identify the most difficult sentences and analyze their characteristics. We employ two ways of defining sentence difficulty. The first one is binary and labels a sentence as difficult if the classifiers fail to correctly predict the sentiment polarity. The second one is a six-level scale based on how many of the top five best-performing classifiers can correctly predict the sentiment polarity. We also define 9 linguistic features  that, combined, aim at estimating the difficulty at sentence level.
\end{abstract}
% IEEEtran.cls defaults to using nonbold math in the Abstract.
% This preserves the distinction between vectors and scalars. However,
% if the journal you are submitting to favors bold math in the abstract,
% then you can use LaTeX's standard command \boldmath at the very start
% of the abstract to achieve this. Many IEEE journals frown on math
% in the abstract anyway.

% Note that keywords are not normally used for peerreview papers.

% For peer review papers, you can put extra information on the cover
% page as needed:
% \ifCLASSOPTIONpeerreview
% \begin{center} \bfseries EDICS Category: 3-BBND \end{center}
% \fi
%
% For peerreview papers, this IEEEtran command inserts a page break and
% creates the second title. It will be ignored for other modes.
\IEEEpeerreviewmaketitle

\section{Introduction}One of the applications of natural language processing is sentiment analysis, which is also known as opinion mining. This field deals with the automatic detection and extraction of subjective information from texts, such as emotions, attitudes, preferences, and opinions.
Sentiment analysis has numerous applications, ranging from marketing to politics, and it has become an increasingly popular topic of research in the past decade \cite{doi:10.1080/19312458.2020.1869198,wankhade2022survey}. Sentiment analysis can be used to identify the sentiments of customers towards a particular product, the opinions of voters towards a political candidate, or the emotions of patients towards their medical condition, among other applications.

 One way to perform sentiment analysis is to examine different levels of granularity: the whole document, a single paragraph, a sentence, or even a specific aspect. However, each level of analysis may encounter the challenges that are discussed previously in this introduction. In this work, we will focus on the most fine-grained level, the aspect level. Indeed, apsect-based is the level for which research is currently the most productive, and consequently also generates the creation of corpora whose expressiveness of sentiment is more subtle and therefore potentially more difficult to analyze. In this article, we will focus on aspect-based sentiment analysis (ABSA).

Despite recent advancements in sentiment analysis, the detection and analysis of sentiments remain challenging due to several factors. One of the most significant challenges in sentiment analysis is the ambiguity of language \cite{cambria2013knowledge,10.1145/3046684,gref-etal-2022-study}. For example, the sentence "The pizza was terrible" is ambiguous and requires more context to be correctly understood. Ambiguity can also be expressed in sentences expressing sentiment through the use of irony and sarcasm. The latter will use terms that convey sentiment, expressing the opposite of the meaning of the sentiment usually used.

Another challenge in sentiment analysis is the detection of tone and sarcasm \cite{maynard2014cares,farias2017irony,li2023detecting}. As we mentioned previously, the detection of tone, sarcasm and irony remains complex to take into account when analyzing sentiments, particularly because of the ambiguity they convey. For instance, the sentence "I just love spending hours on hold with customer service", which uses the word love, which normally conveys a high positive sentiment, expresses in fact a negative sentiment.

Cultural and contextual differences also pose a challenge to sentiment analysis \cite{KONG2023103374}. Sentiments can vary based on culture and context, and what might be considered positive in one culture may not be the same in another. For example, in some cultures, being direct and blunt is considered a positive trait, while in others, it may be seen as negative. In the sentence "That dress is very revealing," the sentiment may be negative in some cultures because revealing clothes are seen as inappropriate or provocative. In other cultures, the sentiment may be positive because revealing clothes are associated with beauty or confidence. Sentiment analysis algorithms must be trained on diverse datasets to overcome such differences.

The diversity of domains within the sentences analyzed can make sentiment analysis more complex. For example, the sentence "The product is lightweight", depending on the domain, may express a positive or a negative sentiment. So if the domain is the sale of laptops, the sentiment will be rather positive, in fact, it implies that the laptop is easy to carry around. On the other hand, if you're selling body-building products, this same sentence may express a rather negative sentiment: it implies that the product might not be heavy enough for a good workout. 

 In addition, the syntactic diversity of sentences can also make analysis more difficult. In particular, if sentences with a more subtle expression of sentiment are used. For instance, the sentence "The politician's actions in the public sphere have not gone unnoticed, hinting at a promising potential for positive transformation in our community" expresses feelings more subtly than in a review such as the sentence "The food was very good.

Additionally, data quality issues such as noise, missing data, or bias \cite{asyrofi2021biasfinder} can affect the accuracy of sentiment analysis. So, for example, the sentence "The film produced by X, it's great" can be positive about the film but also about the production company. Without additional information, it is not possible to analyze the sentence correctly.
  
However, recent years have seen advances in language models, particularly the emergence of BERT \cite{DBLP:conf/naacl/DevlinCLT19} and GPT \cite{NEURIPS2020_1457c0d6}. These have enabled algorithms to better capture the semantics of texts, resulting in a marked improvement in performance. This raises the question of whether the challenges mentioned above still exist, and if so, how algorithms manage to overcome them. In this article, we explore how algorithms handle these difficulties and which subjective sentences pose the greatest challenge.

 Contrary to the usual focus in current aspect-based sentiment analysis research, our aim does not involve achieving better results or introducing a new classification model. Rather, it is to comprehend why existing models are not working in some cases and why some data sets are "simpler" than others. The aim is therefore to observe the behavior of classification models in the aspect-based sentiment analysis task and to estimate the degree of difficulty of the analyzed sentences. Thus, estimating the difficulty of sentiment analysis could enhance performance while minimizing resource usage.

 In order to better understand the aspect-based sentiment analysis difficulty, we raise the following research questions: 

 To summarize, in order to answer these three research questions, we propose in our work to:
\begin{itemize}
     \item What impact does domain and syntactic diversity have on the difficulty of the aspect based sentiment analysis?

     \item Which linguistic features should be chosen to estimate sentence difficulty in aspect-based sentiment analysis?

     \item What is the difference in performance between traditional models and large language models?
\end{itemize}

 To summarize, in order to answer these three research questions, we propose in our work to:
\begin{itemize}
\item Merge three different corpora into a single corpus in order to measure the behavior of models with thematic and syntactic diversity;
\item select 21 models, two different text representations and one LLM in order to analyze their respective behavior and performance when faced with aspect-based sentiment analysis;
\item conduct numerous experiments in order to better understand the challenges faced by the models;
\item Select seven linguistic features to estimate sentence difficulty.
\end{itemize}

 The remainder of this paper is organized as follows. Section \ref{sec:related} provides an overview of aspect-based sentiment analysis and the concept of difficulty in Information Retrieval. Section \ref{sec:corpora} presents 3 different data sets used in the experiments. Section \ref{sec:experiments} explains the experimental protocol and presents the  aspect-based sentiment polarity classification results. Section \ref{sec:difficulty} presents the difficulty definitions and tests if the difficulty may be predicted. Finally, Section \ref{sec:conclusion} concludes the paper and suggests directions for future work.

\section{Related Work}
\label{sec:related}
We divide our related work into themes that influence our analysis, including Aspect-based Sentiment Analysis and Query Difficulty in Information Retrieval.

% needed in second column of first page if using \IEEEpubid
%\IEEEpubidadjcol

\subsection{Aspect-based sentiment analysis}
Sentiment analysis aims to automatically detect and classify the emotions and sentiments that are expressed in written text. There are different levels of granularity for sentiment analysis, such as document level, paragraph level, sentence level, and aspect level, which is the finest level that focuses on the element of subjectivity. The research on sentiment analysis can be categorized into three main approaches: machine learning, deeplearning, and ensemble learning. Some of the most effective machine learning techniques for this task are naive Bayes \cite{villavicencio2021twitter,10.1063/1.4994463,7877424} and SVM \cite{10.1007/978-981-19-8825-7_11,ahmad2018sentiment,fikri2019comparative}. Some of the deeplearning algorithms \cite{LI2020102290} that are used for this task are RNNs \cite{wang2016combination,basiri2021abcdm}, LSTMs \cite{ma2018sentic,rehman2019hybrid,10.1007/978-981-33-4673-4_16} and transformers \cite{hoang2019aspect,gao2019target}. Ensemble-based methods \cite{tiwari_systematic_2023} combine multiple classifiers, which can belong to either of the previous approaches. Due to the large amount of research on sentiment analysis, several surveys have been conducted recently (\cite{8746210,bordoloi_sentiment_2023,cui_survey_2023, tiwari_systematic_2023}). However, in recent years, the research has shifted more towards multimodal, multilingual sentiment analysis and the aspect-based level of sentiment analysis. In this document, we will only focus on aspect-based sentiment analysis. This research on aspect-based sentiment analysis was pioneered by the seminal work of \cite{10.1145/1014052.1014073} and has been developed with the creation of numerous models and data sets in various languages. As in the broader context of sentiment analysis, we find similar algorithms but adapted to the task. Among these algorithms, we can mention the use of SVMs \cite{6637416} and Naïve Bayes \cite{mubarok2017aspect}. More recently, the emergence of deeplearning has significantly improved model performance. For example, there are models using RNNs, LSTMs \cite{ma2018targeted,do2019deep,liu2020aspect} and transformers \cite{Karimi2020ImprovingBP,mutlu_dataset_2022}. For further reading, one can also refer to surveys on the topic (\cite{9996141,10.1145/3503044,chauhan_aspect_2023}).

\subsection{Difficulty in Information Retrieval}
We have seen, in the introduction, that aspect-based sentiment analysis is a challenging task that requires different approaches depending on the difficulty of the sentences. To avoid applying costly algorithms to sentences that are hard to classify, we need a way to identify them and use alternative methods. This saves money and reduces the energy required in order to run the models. Difficulty detection is an important research topic, especially for information retrieval. Many studies have proposed predictors of difficulty based on different criteria, such as distribution \cite{LoupyEvaluationOD}, ambiguity \cite{mothe:halshs-00287692}, and complexity \cite{10.1145/2600428.2609496}. 
In information retrieval, the models that predict difficulty can be classified into two groups: those that use pre-retrieval predictors and those that use post-retrieval predictors. Pre-retrieval predictors are based on statistics of the query terms, as in \cite{10.1007/978-3-540-78646-7_8,cronen2004language,https://doi.org/10.1002/asi.20011}. Post-retrieval predictors rely on the output of the information retrieval models, as in \cite{10.1145/1835449.1835683,10.1145/564376.564429,10.1145/2180868.2180873,10.1145/1277741.1277835,10.1145/2661829.2661906}. With the development of deep neural networks, some recent works have used them to predict difficulty, such as \cite{10.1145/2661829.2661906,10.1145/3341981.3344249}. However, \cite{10.1007/978-3-031-28244-7_15,10.1145/3578337.3605142} have questioned the validity of using difficulty evaluation for Neural Information Retrieval based on PLM (Pre-trained Language Models).

\section{Reference Data sets used for Corpus Building}
\label{sec:corpora}
We selected 3 data sets (Laptops, Restaurant and MTSC) whose purpose is to perform aspect-based classification. The data sets have been created at a 6-7 year time distance, and choosing three data sets that span over such a large period of time would reflect the evolution of the field
The SemEval Laptop Reviews data set \cite{pontiki2014semeval} is often associated with the laptops data set for aspect-based sentiment analysis. It was first used in the SemEval-2014 Task 4: Aspect-Based Sentiment Analysis challenge. This data set is used for two subtasks: aspect identification (Subtask 1) and aspect-oriented sentiment polarity classification (Subtask 2). 

The restaurant data set for aspect-based sentiment analysis consists of more than 3000 English sentences from restaurant reviews initially proposed by Ganu \textit{et al.} \cite{ganu2009beyond}. It has been modified for SemEval-2014 \cite{pontiki2014semeval} to include annotations for aspect terms occurring in the sentences (Subtask 1), aspect term polarities (Subtask 2), and aspect category-specific polarities (Subtask 4). Some errors in the original data set have been corrected. Human annotators identified the aspect terms of the sentences and their polarities (Subtasks 1 and 2). Additional restaurant reviews, not present in the original data set, have been annotated in the same manner and kept by the organizers of SemEval-2014 as test data.

NewsMTSC is an aspect-based sentiment analysis data set proposed by the authors of \cite{Hamborg2021b}. It focuses on news articles about policy issues and contains over 11,000 labeled sentences sampled from online US news outlets. 

Most of the sentences contain several aspects that are mentioned in the data set. The conflict class does not exist, so only three polarity levels may be encountered: "positive", "negative", and "neutral".

The statistics of the data sets used for the experiments are shown in Table~\ref{tab:data_sets}.

\begin{table}[!ht]
\centering
\caption{Summary of the data set information}
\label{tab:data_sets}
\begin{tabular}{lcrrc}
\toprule
\textbf{Data Sets} & \textbf{Total} & \textbf{Train} & \textbf{Test} & \textbf{\# of classes} \\
\midrule
Laptops & 2407 & 2358 & 49 & 4 \\
Restaurants & 3789 & 3693 & 96 & 4 \\
MTSC & 9885 & 8739 & 1146 & 3 \\
\bottomrule
\end{tabular}
\end{table}

\begin{figure*}[!ht]
\centering
\includegraphics[width=\textwidth]{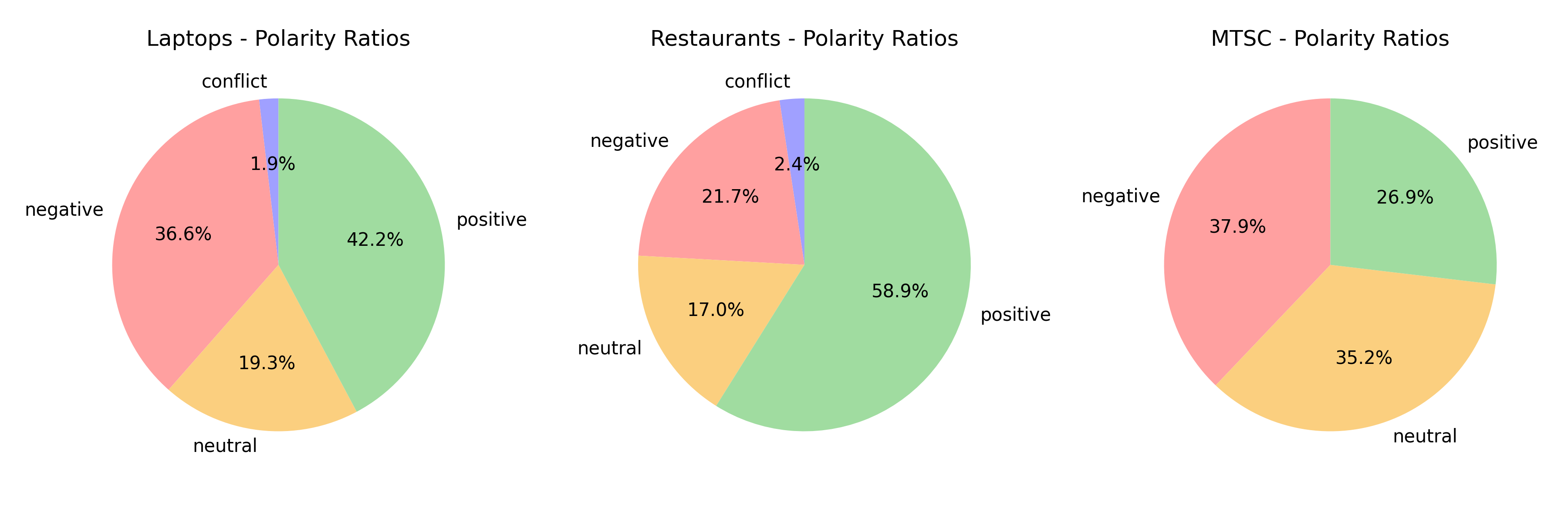}
\caption{Polarity distributions for the three data sets}
\label{fig:polarity_distribution}
\end{figure*}

To compare the polarity distributions in each data set, we plot them in Figure \ref{fig:polarity_distribution}. The data set with the most balanced polarity classes is "MTSC", where $37.9\%$ of the instances are negative. The other two data sets have more skewed distributions, with "Restaurants" having the highest imbalance, with $58.9\%$ of the instances being positive and only $2.4\%$ being conflict. The conflict class is the least frequent in all three data sets, with $1.9\%$ in "Laptops" and zero in "MTSC".

In this section, we present the statistics of the tokens and sentences in the data sets. Table \ref{tab:tokens_sentences} shows the summary of the results. The data set may contain repeated sentences, because one sentence can have multiple aspects and thus be assigned to different categories. However, the number of distinct aspects is smaller than the number of distinct sentences, which implies that some aspects are expressed by different sentences. Moreover, the maximum number of tokens per aspect varies from 6 to 31 on the Laptops and MTSC data sets, respectively. This suggests that aspects can be long and not just consist of one or two words.

\begin{table*}[!ht]
\centering
\caption{Token and sentence statistics for the data sets.}
\label{tab:tokens_sentences}
\begin{tabular}{p{2cm}p{2cm}p{2cm}p{2cm}p{2.5cm}}
\toprule
 \textbf{Data set} & \textbf{\# of observations} & \textbf{\# of unique aspects} & \textbf{\# of unique sentences} & \textbf{Max \# of tokens per aspect} \\
\midrule
    Laptops &           2407 &             1044 &               1484 &                          6 \\
Restaurants &           3789 &             1289 &               2022 &                         19 \\
       MTSC &           9885 &             3525 &               8802 &                         31 \\
\bottomrule
\end{tabular}
\end{table*}

We performed a linguistic analysis of sentiment polarity for each data set to better comprehend the data sets. We computed the mean number of tokens, nouns, verbs, named entities, and adjectives per instance. Table~\ref{tab:linguistic} displays the results. The sentences from MTSC are typically longer than the sentences from the other data set, as one can observe. However, for query difficulty prediction, the query length does not have a correlation with the Average Precision performance measure, as shown by \cite{he2004inferring}. Another noteworthy observation is that the MTSC corpus has a considerably higher mean number of named entities than the other two data sets. This may increase the challenge of aspect-based polarity classification.

\begin{table*}[!ht]
\caption{Average Number of Tokens, Nouns, Verbs, Named Entities, and Adjectives per Sentence}
\label{tab:linguistic}
\centering
\begin{tabular}{p{2.3cm}cccccc}
\toprule
\multicolumn{2}{c}{\textbf{Data set/Class}} & \textbf{Tokens} & \textbf{Nouns} & \textbf{Verbs} & \textbf{Named Entities} & \textbf{Adjectives} \\
\midrule
\multirow{4}{*}{Laptops} & Positive & 20.04 & 3.72 & 1.98 & 0.64 & 1.94 \\
 & Negative & 22.76 & 4.04 & 2.81 & 0.82 & 1.42 \\
 & Neutral & 25.24 & 4.54 & 2.83 & 1.33 & 1.43 \\
 & Conflict & 23.84 & 3.82 & 2.69 & 0.84 & 2.02 \\
\hline
\multirow{4}{*}{Restaurants} & Positive & 18.81 & 3.77 & 1.41 & 0.54 & 2.20 \\
 & Negative & 22.50 & 4.10 & 2.18 & 0.50 & 1.99 \\
 & Neutral & 21.62 & 4.14 & 2.26 & 0.74 & 1.48 \\
 & Conflict & 22.31 & 3.60 & 1.76 & 0.48 & 2.67 \\
\hline
\multirow{4}{*}{MTSC} & Positive & 30.12 & 5.07 & 3.29 & 2.95 & 1.99 \\
 & Negative & 31.58 & 5.31 & 3.55 & 3.27 & 1.94 \\
 & Neutral & 27.63 & 4.13 & 3.03 & 3.30 & 1.40 \\
 & Conflict & 0.00 & 0.00 & 0.00 & 0.00 & 0.00 \\
\bottomrule
\end{tabular}
\end{table*}

\section{Experiments and Results}
\label{sec:experiments}
Firstly, we carried out experiments to better understand the notion of difficulty for learning algorithms for the aspect-based sentiment analysis task.

We conducted experiments with 21 learning models on two different representations (TF-IDF and BERT) and each of these corpora, to find out which corpus was the most challenging for the models. We define a model as challenging if its performance is lower than the median of model performance for a given corpus. We also examine the challenge at a finer level, by identifying the sentences and paragraphs that were the most problematic for the selected models. We have used the models proposed by the LazyText python module (https://github.com/jdvala/lazytext)  for text classification tasks. The experiments were carried out on a server with 40 Intel(R) Xeon(R) CPU at 2.20GHz, 354 Gb and 8 GPU (7 Nvidia A40 and 1 Nvidia GeForce RTX 3090).

The results of these experiments are presented in aggregate form in this article in order to highlight the salient features and to contribute to the discussion on the notion of difficulty for the aspect-based sentiment analysis task.

\begin{table*}[!ht]
\caption{Macro and weighted Metrics of best classification models with TF-IDF and BERT text representation on the Laptops corpus}
\label{tab:laptop}
\centering
\begin{tabular}{cccc}
\toprule
Model                  & Representation & F1(Macro) & F1(Weighted) \\
\midrule
CalibratedClassifierCV & TF-IDF         & \textbf{0.913765}  & 0.937463     \\
DecisionTreeClassifier & TF-IDF         & 0.741935  & \textbf{0.989467}     \\
MLPClassifier          & BERT           & \textbf{0.983598}  & 0.979436     \\
BaggingClassifier      & BERT           & 0.741935  & \textbf{0.989467}  \\
\bottomrule
\end{tabular}
\end{table*}

The table \ref{tab:laptop} presents the best results for the Laptops corpus among the 42 experiments (the 21 models + the 2 representations). We can see that overall, the results reach very high scores. These scores show that it is relatively easy to carry out sentiment analysis on this corpus. However, we note a certain diversity in the results and there are no models that clearly stand out as being the best. Indeed, when a model is the best for Measure F1(Macro), it has weaker results than another model in the case of Measure F1(Weighted). In the experiments, several models achieved the same results. The two best models for the F1(Weighted) measure obtained exactly the same results for this measure as for the F1(Macro) measure. The MLPClassifier model obtained the best overall results.

\begin{table*}[!ht]
\caption{Macro and weighted Metrics of best classification models with TF-IDF and BERT text representation on the Restaurant corpus}
\label{tab:restaurant}
\centering
\begin{tabular}{cccc}
\toprule
Model                  & Representation & F1(Macro) & F1(Weighted) \\
\midrule
DecisionTreeClassifier & TF-IDF         & 1.000000  & 1.000000     \\
BaggingClassifier      & BERT           & 1.000000  & 1.000000     \\
\bottomrule
\end{tabular}
\end{table*}

The table \ref{tab:laptop} presents the best results for the Restaurant corpus among the 42 experiments (the 21 models + the 2 representations). We can see immediately from the results for this corpus, whatever the representation (TF-IDF or BERT) and whatever the measure used (F1(Macro) and F1(Weighted). That the results are identical and obtain the maximum value. In view of the results, this corpus seems easier than the Laptops corpus. Among the models, for the TF-IDF representation, 5 models obtained this maximum score. In the case of the BERT representation, 5 models also obtained the score of 100\%. Among these models, 4 obtained this score for both TF-IDF and BERT. 

\begin{table*}[!ht]
\caption{Macro and weighted Metrics of best classification models with TF-IDF and BERT text representation on the MTSC corpus}
\label{tab:mtsc}
\centering
\begin{tabular}{cccc}
\toprule
Model                  & Representation & F1(Macro) & F1(Weighted) \\
\midrule
BernoulliNB                  & TF-IDF    & 0.613077  & 0.626609     \\
LogisticRegressionCV         & BERT      & 0.708159  & 0.721086     \\
\bottomrule
\end{tabular}
\end{table*}

The table \ref{tab:mtsc} presents the best results for the MTSC corpus among the 42 experiments (the 21 models + the 2 representations). The two models presented in the table both obtain the best scores in F1(Macro) and F1(weighted). We can see that the corpus seems to be much more difficult than the two previous ones. Indeed, the best score drops by more than 20 points compared to the results observed for the Laptops and Restaurants corpora, all representations combined. In addition, the greater difficulty benefits the BERT representation, which, thanks to its greater expressiveness and the fact that it takes semantics into account, enables the models to perform better. The best model using the BERT representation obtained a performance score more than 10 points higher than the best model using the TF-IDF representation.

By comparing the results for the three corpora, we can see that the difficulty has changed. This evolution of difficulty is correlated with the temporal evolution of the corpora. The easiest corpus (restaurant) is the oldest corpus and the most difficult corpus (MTSC) is the most recent corpus. Although each corpus is 6 or 7 years apart, the difficulty seems to have increased more significantly in recent years, in line with the increasing capacity of the models at the same time.

\begin{table*}[!ht]
\caption{Classification results for fine-tuned BERT on Restaurant, Laptops and MTSC}
\label{tab:finetuned}
\centering
\begin{tabular}{ccccc}
\toprule
Measure                  & Corpus & Presision & Recal & F1-Score \\
\midrule
Weighted Avg      & Laptops    & 0.88  & 0.95 & 0.90     \\
Macro Avg         & Laptops    & 0.95  & 0.94 & 0.94   \\
Weighted Avg      & Restaurant & 0.63  & 0.55 & 0.57     \\
Macro Avg         & Restaurant & 0.89  & 0.86 & 0.86   \\
Weighted Avg      & MTSC       & 0.74  & 0.74 & 0.73     \\
Macro Avg         & MTSC       & 0.77  & 0.73 & 0.74   \\
\bottomrule
\end{tabular}
\end{table*}

As models and text representations have evolved, we have added experiments using an LLM. BERT was chosen because of its impressive performance in classification tasks.

Table \ref{tab:finetuned} presents the results obtained using BERT and fine-tuning it for each of the corpora. We can see from the results that the fine-tuned version of BERT has more difficulty analysing sentiment in the case of the Laptops corpus and especially the Restaurant corpus. In fact, the results are 43 points lower when using the F1(Weighted) performance measure and 14 points lower when using the F1(Macro) measure. These disappointing performances are probably due to the size of the neural network used by BERT, which requires much more data. When the data is larger, as is the case in the MTSC corpus, BERT achieves performances similar to those observed among the best previous models. These results confirm our view that it is not always necessary to use models which, although very efficient, require a great deal of resources.

However, the corpora are relatively homogeneous in their structure and domain of application. The Laptops corpus is used to analyse sentiment for reviews of laptops. The Restaurant corpus, as its name clearly indicates, is used to analyse sentiments about restaurants. Finally, the MTSC corpus analyses sentiment in newspaper texts dealing with politics. The syntactic structure of the sentences is also relatively homogeneous in each of the 3 different corpora. Following on from this work, we wanted to analyse whether the semantic and syntactic heterogeneity of the corpora had an impact on the difficulty of aspect-based sentiment analysis. We therefore combined the three initial corpora into a single corpus. In this way, we introduce semantic diversity, since three different domains are represented in this new corpus, but also syntactic diversity, since the Restaurant corpus and the Laptops corpus come from reviews, unlike the MTSC corpus, which comes from news.

% tableau à mettre pour tester la disparité de domaine
\begin{table*}[!ht]
\centering
\caption{Macro Metrics of Classification Models, BERT, aspect at the end, all 3 collections together.}
\label{tab:all_macro}
\begin{tabular}{lrrr}
\toprule
 Model & Precision (Macro) & Recall (Macro) & F1 (Macro) \\
\midrule
AdaBoostClassifier & 0.377904 & 0.345437 & 0.336137 \\
BaggingClassifier & 0.476424 & 0.466165 & 0.468056 \\
BernoulliNB & 0.458911 & 0.396189 & 0.389916 \\
CalibratedClassifierCV & 0.713334 & 0.711479 & 0.710975 \\
DecisionTreeClassifier & 0.419183 & 0.421055 & 0.419640 \\
DummyClassifier & 0.092693 & 0.333333 & 0.145051 \\
ExtraTreeClassifier & 0.374319 & 0.376414 & 0.374709 \\
ExtraTreesClassifier & 0.516363 & 0.507678 & 0.509121 \\
GradientBoostingClassifier & 0.525921 & 0.522387 & 0.523971 \\
KNeighborsClassifier & 0.613998 & 0.596456 & 0.598831 \\
LinearSVC & 0.540271 & 0.533929 & 0.536354 \\
LogisticRegression & 0.734716 & \bfseries 0.727424 & \bfseries 0.730195 \\
LogisticRegressionCV & 0.732542 & 0.724818 & 0.727568 \\
MLPClassifier & 0.533977 & 0.531156 & 0.532430 \\
NearestCentroid & 0.385196 & 0.347173 & 0.344397 \\
PassiveAggressiveClassifier & \bfseries 0.735536 & 0.716113 & 0.721160 \\
Perceptron & 0.682906 & 0.637421 & 0.596026 \\
RandomForestClassifier & 0.529846 & 0.524587 & 0.526170 \\
RidgeClassifier & 0.718174 & 0.712822 & 0.714928 \\
SGDClassifier & 0.734330 & 0.667922 & 0.669870 \\
SVC & 0.723458 & 0.715283 & 0.717820 \\
\bottomrule
\end{tabular}
\end{table*}

Table \ref{tab:all_macro} shows the results of applying the 21 models using the BERT representation for text. The measures used are Macro Precision, Macro Recall and Macro F1. As we can see from analyzing the results, performance varies greatly from one model to another. For all the models, the results are lower than those observed for the Restaurant and Laptops corpora alone. This corpus, aggregated from the three initial corpora, seems more difficult for the models to grasp. Contrary to what we had thought when we launched these experiments, the best models did not have difficulty overcoming the multi-domain or the syntactic diversity of the sentences from the different corpora. On the contrary, for some models, which were already obtaining the best results with the "monocorpus", the results were better than those obtained with the MTSC corpus alone. This finding implies that performance is certainly degraded compared to the models trained on the restaurant and laptop corpora alone, and that consequently the most difficult corpus 'takes precedence' over the easier corpora, albeit with a slight improvement in results in this case. This slight improvement in results compared to the MTSC treatment alone is due to the addition of new sentences that are easier to classify. 

% tableau à mettre pour tester la disparité de domaine

\begin{table*}[!ht]
\centering
\caption{Weighted Metrics of Classification Models, BERT, aspect at the end, all 3 collections together.}
\label{tab:all_weighted}
\begin{tabular}{lrrr}
\toprule
 Model & Precision (Weighted) & Recall (Weighted) & F1 (Weighted) \\
\midrule
AdaBoostClassifier & 0.492801 & 0.463207 & 0.442574 \\
BaggingClassifier & 0.628818 & 0.622773 & 0.621522 \\
BernoulliNB & 0.609801 & 0.558482 & 0.541243 \\
CalibratedClassifierCV & 0.713001 & 0.712626 & 0.711282 \\
DecisionTreeClassifier & 0.559612 & 0.557707 & 0.558077 \\
DummyClassifier & 0.077328 & 0.278079 & 0.121007 \\
ExtraTreeClassifier & 0.499498 & 0.498064 & 0.497937 \\
ExtraTreesClassifier & 0.684477 & 0.680868 & 0.678884 \\
GradientBoostingClassifier & 0.701288 & 0.699458 & 0.700155 \\
KNeighborsClassifier & 0.609640 & 0.601859 & 0.599534 \\
LinearSVC & 0.718596 & 0.715724 & 0.716243 \\
LogisticRegression & \bfseries 0.732239 & \bfseries 0.730442 & \bfseries 0.730539 \\
LogisticRegressionCV & 0.731026 & 0.729667 & 0.729344 \\
MLPClassifier & 0.709147 & 0.707978 & 0.708401 \\
NearestCentroid & 0.521419 & 0.491092 & 0.481640 \\
PassiveAggressiveClassifier & 0.730651 & 0.724245 & 0.723283 \\
Perceptron & 0.686317 & 0.628195 & 0.587748 \\
RandomForestClassifier & 0.703830 & 0.701782 & 0.701443 \\
RidgeClassifier & 0.716414 & 0.715724 & 0.715564 \\
SGDClassifier & 0.721423 & 0.687839 & 0.677552 \\
SVC & 0.721415 & 0.720372 & 0.719490 \\
\bottomrule
\end{tabular}
\end{table*}

The Table \ref{tab:all_weighted} shows the results of applying the 21 models using the BERT representation for text. The measures used are Weighted Precision, Weighted Recall and Weighted F1. The findings are similar to those observed with the Macro measures. The results remain very close, for the best models, to those observed when using the MTSC corpus alone, with a very slight increase. In this case, therefore, there was no increase in difficulty through the addition of new domains or through syntactic differentiation of the added sentences. The most difficult corpus takes precedence over the easiest corpora and the addition of new vocabularies or new examples neither increases nor degrades the performance observed initially.

\begin{table}[!ht]
  \centering
  \caption{BERT fine-tuned all collections.}
    \begin{tabular}{lrrr}
    \hline
          & Positive & Negative & Neutral \\
    \hline
    Positive & 284   & 14    & 61 \\
    Negative & 32    & 353   & 78 \\
    Neutral & 45    & 21    & 403 \\
    \hline
    \end{tabular}%
  \label{tab:all_finetune_confusion}%
\end{table}%

\begin{table}[!ht]
  \centering
  \caption{Classification Report: BERT fine-tuned, all collections.}
    \begin{tabular}{lrrrr}
    \toprule
          & Precision & Recall & F1-score & Support \\
    \midrule
    Positive     & 0.79    & 0.79   & 0.79     & 359 \\
    Negative     & 0.91    & 0.76   & 0.83     & 463 \\
    Neutral      & 0.74    & 0.86   & 0.80     & 469 \\
    Macro Avg    & 0.81    & 0.80   & 0.81     & 1291 \\
    Weighted Avg & 0.82    & 0.81   & 0.81     & 1291 \\
    \bottomrule
    \end{tabular}%
  \label{tab:all_finetune}%
\end{table}%

Tables \ref{tab:all_finetune} and \ref{tab:all_finetune_confusion} present the results of using BERT fine-tuned model applied to the three merged corpora. The results are clearly better than those obtained using the 21 models based on BERT textual representation. Referring to the F1(Weighted) and F1(Macro) measures, the results are 8 points better than the best of these models. The results are also better than those obtained using BERT Fine-tuned for the MTSC corpus. On the other hand, the results remain well below the best models used on the Restaurant and Laptops corpora alone. These experiments show the ability of fine-tuned BERT to generalize even when confronted, as is the case here, with several domains and several syntactic forms of sentences when it has sufficient data. However, the difficulty associated with MTSC remains and prevents the model from obtaining higher performance scores.

Through these experiments we were able to see the different levels of difficulty of the corpora. The MTSC corpus is more difficult than the Restaurant and Laptops corpora. When the corpora are easy, it is not necessary to use the most complex models requiring significant resources. On the other hand, when the corpus is more complex, it becomes necessary to use the most advanced models if enough data is available. However, within a corpus, in the case of aspect-based sentiment analysis, complexity can vary from sentence to sentence. The following section defines the notion of complexity, drawing on work carried out in the context of information retrieval.

\section{The notion of difficulty for the ABSA task}
\label{sec:difficulty}
Correctly predicting difficult sentences in the context of aspect-based sentiment analysis can lead to effective selective approaches. We can leave the easy sentences for light models that don't require much computation, and submit the difficult sentences to heavy, complex models. This way, we can achieve a balance between efficiency and effectiveness.
In this section, we give a definition of difficulty for the aspect-based sentiment analysis task. This definition is obtained from the experiments we presented in the previous section. Next, we present experiments on the difficulty estimation. The Restaurant and Laptops corpora were identified as being easy to analyze for the aspect-based sentiment analysis task, so we focus solely on the MTSC corpus in this section.

\subsection{Definition of difficulty}
We propose to explore two definitions of difficulty in this section. The first definition of difficulty is a binary definition. The second definition of difficulty has several levels.

\textbf{Binary definition.} We performed an analysis of the incorrect predictions for the MTSC dataset to evaluate this strategy. We examined the majority votes given by the top five classifiers with both TF-IDF and BERT text representations. We labeled the sentences that were misclassified by both text representations as 'difficult'. For example, a sentence belongs to the difficult class if the majority votes of both the BERT and TF-IDF models classify it incorrectly; otherwise, it belongs to the easy class. On the test set, we discovered 197 in the MTSC data set.

\textbf{Fine-grained definition.} We used different levels of difficulty to implement this strategy, based on how many of the top 5 models correctly classified each sentence, using both text representations. For example, level 0 sentences are the most difficult, because none of the top 5 models got them right. In contrast, level 5 sentences. For BERT representation, the most common classes were level 5, with 693 sentences, and level 0, with 217 sentences. The other 236 sentences were fairly balanced among the other 4 classes.

\subsection{Difficulty prediction}
To make this prediction, we use nine features that we think are relevant for predicting sentence difficulty.

\begin{itemize}
\item Number of nouns: the number of nouns in the sentence
\item Number of Verbs: the number of verbs in the sentence
\item Number of Adjectives: the number of adjectives in the sentence
\item Number of adverbs: the number of adverbs in the sentence
\item Number of named entities: the number of named entities
\item Contains negation: the presence of negation
\item Aspect POS Tag: the aspect POS Tag
\item Average number of synset: the average number of synsets for the aspect
\item Sentence length: the total length of the sentence
\end{itemize}

One can easily notice that both strategies yield unbalanced test data, in terms of class membership. This leads us to consider two types of sampling, the default one, without any class balancing, and the SMOTE sampling \cite{chawla2002smote}, an oversampling technique where the synthetic samples are generated for the minority class. The experiments are carried out using a Kfolds strategy for learning and testing. In these experiments, we selected a $K=10$. We use the 21 classifiers we used previously. The performance measure used in these experiments is the average of the KFold accuracy.

\begin{table}[!ht]
\centering
\caption{Mean scores of classifiers, two levels of difficulty, MTSC test data, linguistic features.}
\label{tab:difficulty_2}
\begin{tabular}{cc}	

\toprule
Classifier & Mean Score\\
\midrule
AdaBoostClassifier & 0.8159\\
BaggingClassifier & 0.7985\\
BernoulliNB & 0.8264\\
CalibratedClassifierCV & 0.8272\\
DecisionTreeClassifier & 0.6841\\
DummyClassifier & \textbf{0.8281}\\
ExtraTreeClassifier & 0.7094\\
ExtraTreesClassifier & 0.8071\\
GradientBoostingClassifier & 0.8133\\
KNeighborsClassifier & 0.8054\\
LinearSVC & 0.7532\\
LogisticRegression & \textbf{0.8281}\\
LogisticRegressionCV & 0.8272\\
MLPClassifier & 0.8246\\
NearestCentroid & 0.5960\\
PassiveAggressiveClassifier & 0.7088\\
Perceptron & 0.6385\\
RandomForestClassifier & 0.8168\\
RidgeClassifier & 0.8272\\
SGDClassifier & 0.8028\\
SVC & \textbf{0.8281}\\
\bottomrule\end{tabular}

\end{table}

The table \ref{tab:difficulty_2} presents the results of the 21 models for the classification of sentences according to two classes: difficult and easy. No resampling is used to compensate for unbalanced classes. Overall, the classifiers don't seem to be able to classify sentences between easy and difficult. The best results are obtained by the SVC and LogisticRegression models. However, the fact that the corpus is unbalanced suggests that the majority class is favored over the other. This hypothesis is verified by the fact that the dummyclassifier obtains the same results as the best classifiers used. As a reminder, the dummyclassifier always predicts the most frequent class. However, if there is a large imbalance between the two classes, this means that the other best classifiers will behave in a similar way.

\begin{table}[!ht]
\centering
\caption{Mean scores of classifiers, two levels of difficulty, MTSC test data, linguistic features, with SMOTE resampling.}
\label{tab:difficulty_2_smote}
\begin{tabular}{cc}\toprule
Classifier & Mean Score\\\midrule
AdaBoostClassifier & 0.7932\\
BaggingClassifier & 0.7635\\
BernoulliNB & 0.5934\\
CalibratedClassifierCV & 0.5864\\
DecisionTreeClassifier & 0.6920\\
DummyClassifier & \textbf{0.8281}\\
ExtraTreeClassifier & 0.6867\\
ExtraTreesClassifier & 0.7871\\
GradientBoostingClassifier & 0.8072\\
KNeighborsClassifier & 0.5209\\
LinearSVC & 0.5658\\
LogisticRegression & 0.5837\\
LogisticRegressionCV & 0.5768\\
MLPClassifier & 0.5986\\
NearestCentroid & 0.5916\\
PassiveAggressiveClassifier & 0.6072\\
Perceptron & 0.5808\\
RandomForestClassifier & 0.7827\\
RidgeClassifier & 0.5838\\
SGDClassifier & 0.3746\\
SVC & 0.6432\\
\bottomrule\end{tabular}

\end{table}

In the case of using SMOTE resampling, we can see in the table \ref{tab:difficulty_2_smote} that the results are not very different from those presented without resampling. The results are even lower overall than those without resampling. In this case, SMOTE resampling doesn't seem to do much to improve performance. The highest performance measure is achieved by the \textit{DummyClassifier}, which shows that the task is not an easy one and that unbalanced data play a major role.

\begin{table}[!ht]
\centering
\caption{Mean scores of classifiers, six levels of difficulty, MTSC test data, linguistic features.}
\label{tab:difficulty_6}
\begin{tabular}{cc}\toprule
Classifier & Mean Score\\\midrule
AdaBoostClassifier & 0.5715\\
BaggingClassifier & 0.4991\\
BernoulliNB & 0.6030\\
CalibratedClassifierCV & \textbf{0.6047}\\
DecisionTreeClassifier & 0.3762\\
DummyClassifier & \textbf{0.6047}\\
ExtraTreeClassifier & 0.3988\\
ExtraTreesClassifier & 0.5462\\
GradientBoostingClassifier & 0.5524\\
KNeighborsClassifier & 0.4957\\
LinearSVC & 0.4834\\
LogisticRegression & 0.6021\\
LogisticRegressionCV & \textbf{0.6047}\\
MLPClassifier & 0.6030\\
NearestCentroid & 0.3419\\
PassiveAggressiveClassifier & 0.4867\\
Perceptron & 0.4033\\
RandomForestClassifier & 0.5716\\
RidgeClassifier & 0.6021\\
SGDClassifier & 0.5151\\
SVC & \textbf{0.6047}\\
\bottomrule\end{tabular}

\end{table}

The second definition of difficulty we use has 6 degrees of difficulty. Table \ref{tab:difficulty_6} shows the results of classifying sentences into these 6 degrees without using SMOTE resampling. We can see that it is very difficult for the models to obtain good results. The best models are CalibratedClassifierCV, LogisticRegressionCV, and SVC with an average accuracy of 0.6047. However, this score is also achieved by DummyClassifier. The small amount of data coupled with an imbalance between classes explains this poor performance. SMOTE resampling was unable to compensate for the unbalanced data.

\begin{table}[!ht]
\centering
\caption{Mean scores of classifiers, six levels of difficulty, MTSC test data, linguistic features, SMOTE resampling.}
\label{tab:difficulty_6_smote}
\begin{tabular}{cc}\toprule
Classifier & Mean Score\\\midrule
AdaBoostClassifier & 0.3351\\
BaggingClassifier & 0.3874\\
BernoulliNB & 0.3141\\
CalibratedClassifierCV & 0.1170\\
DecisionTreeClassifier & 0.3332\\
DummyClassifier & 0.1893\\
ExtraTreeClassifier & 0.3533\\
ExtraTreesClassifier & 0.4799\\
GradientBoostingClassifier & \textbf{0.4826}\\
KNeighborsClassifier & 0.1624\\
LinearSVC & 0.1170\\
LogisticRegression & 0.1318\\
LogisticRegressionCV & 0.1309\\
MLPClassifier & 0.2740\\
NearestCentroid & 0.2842\\
PassiveAggressiveClassifier & 0.1547\\
Perceptron & 0.2702\\
RandomForestClassifier & 0.4799\\
RidgeClassifier & 0.1283\\
SGDClassifier & 0.2504\\
SVC & 0.2436\\
\bottomrule\end{tabular}

\end{table}

The results presented in the table are not conclusive. Indeed, although we use SMOTE resampling, the best model peaks at 0.4826, which is a relatively low value. Although the task is difficult because of the small amount of data and the unbalanced classes, the features selected do not seem to be the most relevant for obtaining the best results for predicting difficulty. The paucity of data suggests that a fewshot method might be a solution to consider in order to obtain better results.

\section{Conclusion}
\label{sec:conclusion}
This paper aims to explore the factors that affect sentence difficulty in aspect-based sentiment analysis, without proposing new models or improving existing ones. This is a novel topic inspired of difficulty analysis in information retrieval. We performed experiments to understand the impact of domain and syntactic diversity on difficulty. We determined nine linguistic features to estimate sentence difficulty in an aspect-based sentiment analysis task. We tried to address the three research questions outlined in the introduction. We do this by examining the results of our experiments and analysis.

\begin{itemize}
\item \textbf{What impact does domain and syntactic diversity have on the difficulty of the aspect based sentiment analysis?} As far as we could see in the experiments, there was little impact from the addition of new domains or syntactic diversity when we compared the results obtained after merging the three corpora with the results observed when using the classic models on the MTSC corpus. However, the results are improved when BERT fine-tuned is used. This improvement is probably due to the increase in training data which allows this model to better fine-tune its parameters on the merged corpus rather than on a single corpus.

\item \textbf{Which linguistic features should be chosen to estimate sentence difficulty in aspect-based sentiment analysis?}
In view of the fairly poor results from the experiments using the nine linguistic features, we can conclude that the features we have selected are not sufficient to successfully analyze sentence-level difficulty.

\item \textbf{What is the difference in performance between traditional models and large language models (LLM)?}
As we have already seen when answering the first research question, LLM results start to outperform the results of more conventional models when the amount of data is sufficient. In particular, BERT fine-tuned significantly outperforms the results of the 21 models when used on the fusion of the three corpora.
\end{itemize}

As future work, we plan to continue studying the notion of difficulty in the aspect-based sentiment analysis task by studying other corpora in other domains with greater difficulty than those of Restaurant and Laptops. We also intend to define other linguistic characteristics in order to better estimate difficulty. We  want to define predictors inspired by Query Performance Prediction (QPP) in IR, as well. We would like to have a more formal definition of the notion of difficulty, based on the experiments that we have already carried out.

% if have a single appendix:
%\appendix[Proof of the Zonklar Equations]
% or
%\appendix  % for no appendix heading
% do not use \section anymore after \appendix, only \section*
% is possibly needed

% use appendices with more than one appendix
% then use \section to start each appendix
% you must declare a \section before using any
% \subsection or using \label (\appendices by itself
% starts a section numbered zero.)
%

% Can use something like this to put references on a page
% by themselves when using endfloat and the captionsoff option.
\ifCLASSOPTIONcaptionsoff
  \newpage
\fi

% trigger a \newpage just before the given reference
% number - used to balance the columns on the last page
% adjust value as needed - may need to be readjusted if
% the document is modified later
%\IEEEtriggeratref{8}
% The "triggered" command can be changed if desired:
%\IEEEtriggercmd{\enlargethispage{-5in}}

% references section

% can use a bibliography generated by BibTeX as a .bbl file
% BibTeX documentation can be easily obtained at:
% http://www.ctan.org/tex-archive/biblio/bibtex/contrib/doc/
% The IEEEtran BibTeX style support page is at:
% http://www.michaelshell.org/tex/ieeetran/bibtex/
\bibliographystyle{IEEEtran}
% argument is your BibTeX string definitions and bibliography database(s)
%\bibliography{IEEEabrv,../bib/paper}
%
% <OR> manually copy in the resultant .bbl file
% set second argument of \begin to the number of references
% (used to reserve space for the reference number labels box)
%\begin{thebibliography}{1}
\bibliography{IEEEabrv,bibli.bib}

%\end{thebibliography}

% biography section
% 
% If you have an EPS/PDF photo (graphicx package needed) extra braces are
% needed around the contents of the optional argument to biography to prevent
% the LaTeX parser from getting confused when it sees the complicated
% \includegraphics command within an optional argument. (You could create
% your own custom macro containing the \includegraphics command to make things
% simpler here.)
%\begin{biography}[{\includegraphics[width=1in,height=1.25in,clip,keepaspectratio]{mshell}}]{Michael Shell}
% or if you just want to reserve a space for a photo:

%\begin{IEEEbiography}{Michael Shell}
%Biography text here.
%\end{IEEEbiography}

% if you will not have a photo at all:
%\begin{IEEEbiographynophoto}{John Doe}
%Biography text here.
%\end{IEEEbiographynophoto}

% insert where needed to balance the two columns on the last page with
% biographies
%\newpage

%\begin{IEEEbiographynophoto}{Jane Doe}
%Biography text here.
%\end{IEEEbiographynophoto}

% You can push biographies down or up by placing
% a \vfill before or after them. The appropriate
% use of \vfill depends on what kind of text is
% on the last page and whether or not the columns
% are being equalized.

%\vfill

% Can be used to pull up biographies so that the bottom of the last one
% is flush with the other column.
%\enlargethispage{-5in}

% that's all folks
\end{document}